\pdfoutput=1

\documentclass{article}

\usepackage{PRIMEarxiv}

\usepackage[utf8]{inputenc} 
\usepackage[T1]{fontenc}    
\usepackage{hyperref}       
\usepackage{url}            
\usepackage{booktabs}       
\usepackage{amsfonts}       
\usepackage{nicefrac}       
\usepackage{microtype}      
\usepackage{lipsum}
\usepackage{fancyhdr}       
\usepackage{graphicx}       
\graphicspath{{media/}}     

\usepackage[ruled, lined, linesnumbered, commentsnumbered, noend]{algorithm2e}
\SetKwComment{Comment}{/* }{ */}

\usepackage{amsmath}
\usepackage{amssymb}

\usepackage{newtxmath}

\usepackage{multirow}
\usepackage{tikz}
\usetikzlibrary{arrows,automata}
\usetikzlibrary{positioning, fit, arrows.meta, shapes}

\usepackage{pgfplots}
\pgfplotsset{compat=newest}
\usepgfplotslibrary{groupplots}
\usepgfplotslibrary{dateplot}
\usepackage{pgf}

\usepackage{subcaption}
\usepackage{xargs}  
\usepackage[colorinlistoftodos,prependcaption,textsize=tiny]{todonotes}

\newtheorem{definition}{Definition}

\title{Human Guided Learning of Transparent Regression Models
}

\author{
  Lukas Pensel, Stefan Kramer \\
  Institut f\"ur Informatik \\
  Johannes Gutenberg-Universit\"at \\
  Mainz\\
  \texttt{pensel@uni-mainz.de,kramer@informatik.uni-mainz.de} \\
}

\begin{document}
\maketitle

\begin{abstract}
We present a human-in-the-loop (HIL) approach to permutation regression, the novel task of predicting a continuous value for a given ordering of items. The model is a gradient boosted regression model that incorporates simple human-understandable constraints of the form x < y, i.e. item x has to be before item y, as binary features.  The approach, HuGuR (\underline{Hu}man \underline{Gu}ided \underline{R}egression), lets a human explore the search space of such transparent regression models. Interacting with HuGuR, users can add, remove, and refine order constraints interactively, while the coefficients are calculated on the fly. We evaluate HuGuR in a user study and compare the performance of user-built models with multiple baselines on 9 data sets. The results show that the user-built models outperform the compared methods on small data sets and in general perform on par with the other methods, while being in principle understandable for humans. On larger datasets from the same domain, machine-induced models begin to outperform the user-built models. Further work will study the trust users have in models when constructed by themselves and how the scheme can be transferred to other pattern domains, such as strings, sequences, trees, or graphs.
\end{abstract}


\section{Introduction}
\label{introduction}

\begin{figure}[t!]
    \centering
\includegraphics[width=1.0\linewidth]{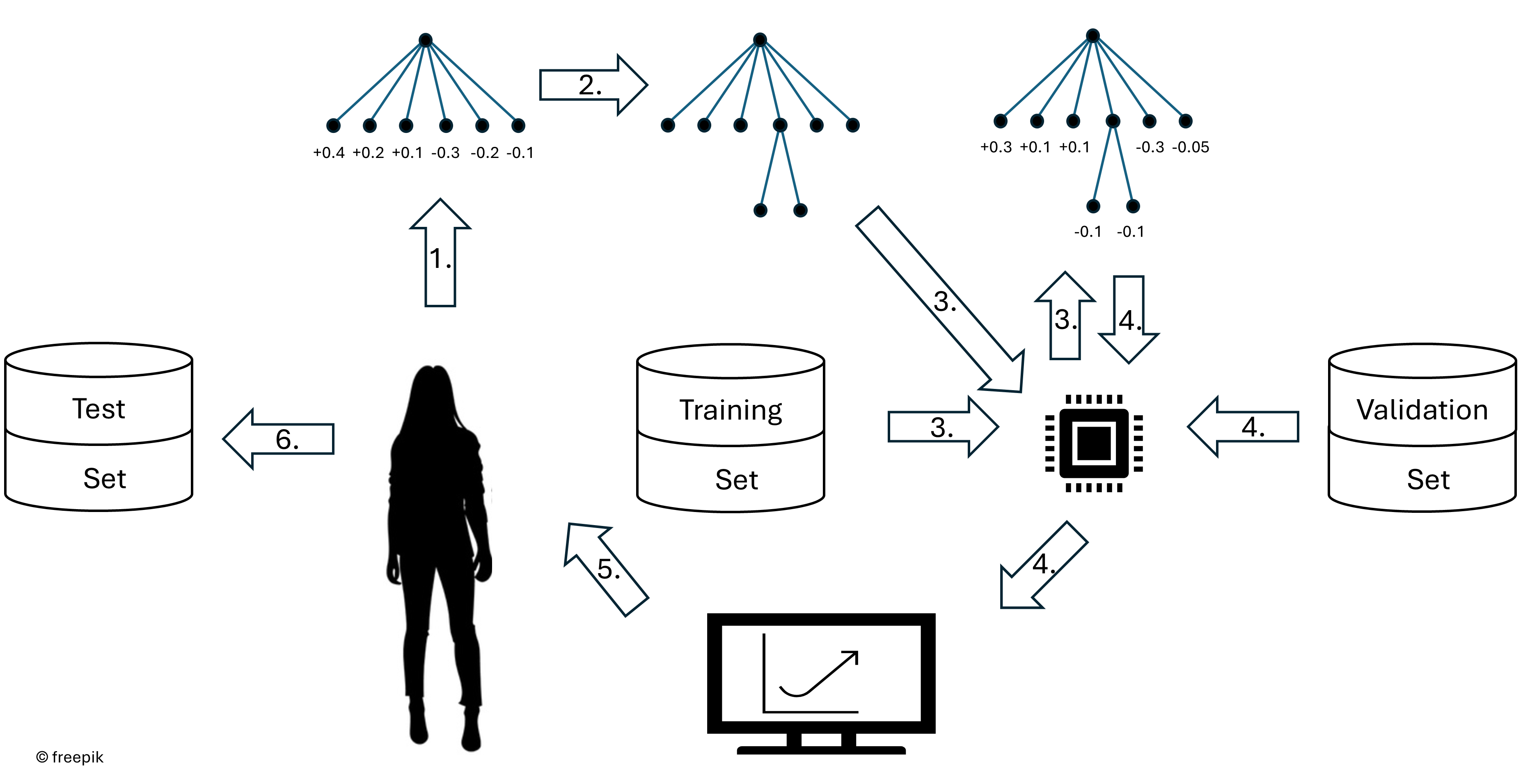}
    \caption{In Human Guided Regression (HuGuR), the user 
    chooses which part of a model to refine (1.), the newly refined model (after step 2.) still has to be parameterized based on the training set (step 3.), the resulting model is validated on the validation set (4.), the performance is monitored and presented to the user (5.). The model refinement goes in cycles. Upon completion, the model chosen as the best is tested on the test set (6.).}
    \label{fig:overview}
\end{figure}

With the steady increase of AI systems that make or support decisions with a potentially high impact on human lives, the need to understand and explain those decisions increases \cite{xai_right}. To achieve this, one can either employ inherently interpretable algorithms, for instance linear models, or incorporate an algorithm that derives explanations and insights from an existing black-box model, for example counterfactuals \cite{counterfactuals},  with the former solution arguably more preferable than the latter in sensitive domains \cite{stop}.

Human-in-the-loop (HIL) machine learning approaches become more and more relevant in machine learning \cite{HILsurvey}, since they allow the incorporation of human knowledge into the learning process. HIL machine learning can be seen as one approach to achieve transparent models. Most HIL machine learning approaches can be categorized into active learning \cite{activelearning}, interactive machine learning \cite{interactivelearning} and machine teaching \cite{machineteaching1}.

The paper presents a new HIL framework called HuGuR (\underline{Hu}man \underline{Gu}ided \underline{R}egression) for building interpretable regression models in domains with structured features. We instantiate the framework for the novel task of what we call {\em permutation regression} in the following: the task of predicting a numerical value for a permutation of a given set of items. This can be applied, for instance, to the problem of predicting the study success of a student solely based on the order of the attended courses, or predicting the performance of a classifier chain\cite{classifierchains}. The task is related to other learning tasks, such as sequence-based prediction -- some reductions are shown in Section 2.3 -- and structured output prediction (SOP) \cite{SOP1}. However, unlike SOP, permutation regression does not require any additional information about the problem besides the permutation itself.
The structured features in our case are constraints on the order of the items. 
The solution (see Figure \ref{fig:overview}) incorporates aspects of transparency and human-understandability coupled with the possibility for the user to interact with the model during its generation. HuGuR combines the transparency of a linear model composed of human-understandable constraints with the possibility for the users to interactively guide the model generation process. In contrast, other interactive machine learning systems \cite{IA1} give the user far less direct control over the model and do not provide a similar level of insight about the model to the user.

In summary, the contributions of this paper are as follows:
\begin{itemize}
    \item[$\bullet$] We introduce the novel task of permutation regression and provide multiple real-world based data sets for it.
    \item[$\bullet$] We present HuGuR, an interactive human-in-the-loop approach based on human-understandable binary features for this task.
    \item[$\bullet$] We conduct a user study to measure HuGuR's performance in its designated field of application. Doing so, we address the research question whether HIL based regression can outperform a pure machine regression, and if so, whether this is dependent on the sample size.
\end{itemize}

The paper is organized as follows: In Section \ref{theory} we define the novel task of permutation regression and introduce our approach to human guided learning of transparent models for permutation regression. In addition, we provide four baseline methods for the task. The real-world based data sets from our experiments are described in Section \ref{data}. The conducted user study and the results are presented in Section \ref{experiments}, before we conclude in Section \ref{conclusion}.
\section{Problem Definition and Methods}
\label{theory}
\subsection{Permutation Regression}

Permutation regression describes the task of estimating the relation between different orderings of a set of items and the respective dependent numerical variable derived by an unknown function. 
\par

\begin{definition}
Given a set of permutations $X$ of the same item set $S$, with $x:S \leftrightarrow S$ being a bijection $\forall x \in X$, and a function $f:X \mapsto Y \subset \mathbb{R}$, \textbf{permutation regression} is the task of finding parameters $\theta \in \Theta$ for a prediction function $f'_{\theta} : X \mapsto \tilde{Y}$ such that it minimizes a given loss function $\mathcal{L}$:
\begin{equation*}
    \min_{\forall \theta \in \Theta} \epsilon_{\theta} = \sum_{x \in X}\mathcal{L}(f(x),f'_{\theta}(x))
\end{equation*} 
\label{def:permutationregression}
\end{definition}

\subsection{Transparent Permutation Regression}
\label{transpermreg}
In order to obtain a transparent model for permutation regression, we base our approach on a gradient boosted regression model on simple binary features, derived from human-understandable constraints, as inputs. Given a set of permutations $X$, we generate a set of $l$ constraints $\rho$ and a function $g_{\rho} : X \mapsto Z \subset \{0,1\}^l$,  which maps each permutation to a binary vector of length $l$ with:
\begin{equation*}
    g_{\rho} : x \rightarrow z \text{ where} \begin{cases}
        z_i = 1&  \text{if } \varrho_i \subset x \\
        z_i = 0& \text{if } \varrho_i \not\subset x
    \end{cases} \forall i \in \{1,\dots,l\}
\end{equation*}
The notation $\varrho_i \subset x$ denotes that the constraint $\varrho_i \in \rho$ is fulfilled by permutation $x$ and analogously the notation $\varrho_i \not\subset x$ denotes that the constraint is not fulfilled.

A constraint $\varrho_i$ is an ordered subset of the permutations' underlying item set $S$. A permutation $x$ fulfills a constraint $\varrho_i$ if the elements of $\varrho_i$ occur in the same order in $x$. For instance, the permutation $x=[ \, 3, 2, 1, 4] \,$ fulfills the constraint $\varrho_i=(2,4)$ but not the constraint $\varrho_j=(1,2,3)$.

The final part of the model are the coefficients $\beta$ for each feature and the baseline prediction $\mu$. We set $\mu$ as the average target value of the training set provided. We calculate the coefficients $\beta$ by minimizing the predictive error $\epsilon = \sum_{i=1}^m |y_i - \tilde{y}_i|$, where $y_i$ is the true target value of permutation $x_i$, and $\tilde{y}_i$ is the predicted target for $x_i$. This yields the linear model:
\begin{equation*}
    \tilde{y}_j = \mu + \sum_{i=1}^{l}\beta_i g_{\rho}(x_j)_i
\end{equation*}

In order to find the constraints that greatly improve the predictive performance of our model, we employ an iterative greedy gradient boosting \cite{gradboost} based approach incorporating a fast Gauss-Southwell selection \cite{gauss_southwell,seq_reg}, similar to related methods for graph \cite{graph_reg} and sequence \cite{seq_reg} regression. Hereby we explore the space of all possible constraints with a breadth-first search, calculate the gradient for each visited constraint and add the constraint with the highest absolute gradient to our constraint set. We prune unpromising branches in the search tree to speed up the search by employing an upper bound, that exploits the hierarchical structure of the constraints similar to the exploitation of the structure of the subsequence space\cite{seq_reg}.

After each iteration, we calculate the coefficient $\beta^*$ for the new constraint $\varrho^*$ using the constraint's gradient $\tau$ and calculate the new residuals $\delta$, where each element $\delta_i = y_i - \tilde{y}_i$ is the difference between the real target $y_i$ and the predicted target $\tilde{y}_i$ of a permutation $x_i$. We continue this iterative process of selecting a constraint and calculating its coefficient until we have generated $l$ constraints, where $l$ is a selectable hyperparameter.

\subsection{Baseline models}

In order to use a permutation of an item set as input for a conventional machine learning algorithm, the permutation has to be transformed. In this section, we introduce multiple possible encodings for permutations. We start with the simplest possible baseline.

\subsubsection{Naive baseline}

This simple model always predicts the average target value over the provided training set.

\subsubsection{One-Hot Encoding}

The first method we use to encode the permutations is a binary one-hot encoding of the items' indices. For each permutation $x = \left[x_1, \dots, x_n\right]$, all integer entries $x_i$ are replaced by a zero-vector with a singular one at position $x_i$, and concatenated to a single vector.

\begin{equation*}
 [ \, 3, 2, 1, 4] \, \Rightarrow [ \,0010\ 0100\ 1000\ 0001] \,
\end{equation*}

\subsubsection{Graph Encoding}

For the directed graph $G = (V,E)$, where $V$ describes the set of all nodes and $E$ describes the set of all edges, each item of the underlying item set corresponds to a node $v \in V$ and each edge $e \in E$ describes a possible succession in the permutation. Since every item can succeed every other item in at least one unique permutation, the graph is complete. Each permutation can be represented by a path in the graph using all nodes. This structure can be used to encode a permutation.

A permutation of $n$ items is encoded in a binary vector of length $(n + 2) \times n$, where the first $n\times n$ values are a one-hot encoding of the edges present in the path representation of the permutation and the next $2n$ values are the one-hot encoding of the first and last item of the permutation.
\begin{equation*}
 [ \, 3, 2, 1, 4] \, \Rightarrow [ \, \underbrace{0001\ 1000\ 0100\ 0000}_{\text{\makebox[0pt][c]{encoding of path edges}}}\ \overbrace{0010\ 0001}^{\text{\makebox[0pt][c]{encoding of start and end}}}] \,
\end{equation*}

In the example, the first block of four binary values represents an edge from 1 to 4, the second block of four binary values an edge from 2 to 1, and so forth.

\subsubsection{Sequence Encoding}
A permutation of $n$ items can be viewed as a sequence of length $n$, where each item appears exactly once in the sequence. Similarly to the first method, we use a binary one-hot encoding, but instead of concatenating the vectors we keep them as a sequence of one-hot encoded items. Therefore, we can incorporate a recurrent neural network architecture, such as long short-term memory (LSTM), to connect the permutation data with a simple multilayer perceptron (MLP). 

\begin{equation*}
 [ \, 3, 2, 1, 4] \, \Rightarrow \begin{bmatrix}
 0010 \\
 0100 \\
 1000 \\
 0001
 \end{bmatrix}
\end{equation*}
\par

\subsection{Human Guided Regression Model}
\label{sec:HIL}

\begin{figure}[!htbp]
    \centering
    \makebox[\textwidth][c]{\includegraphics[width=\linewidth]{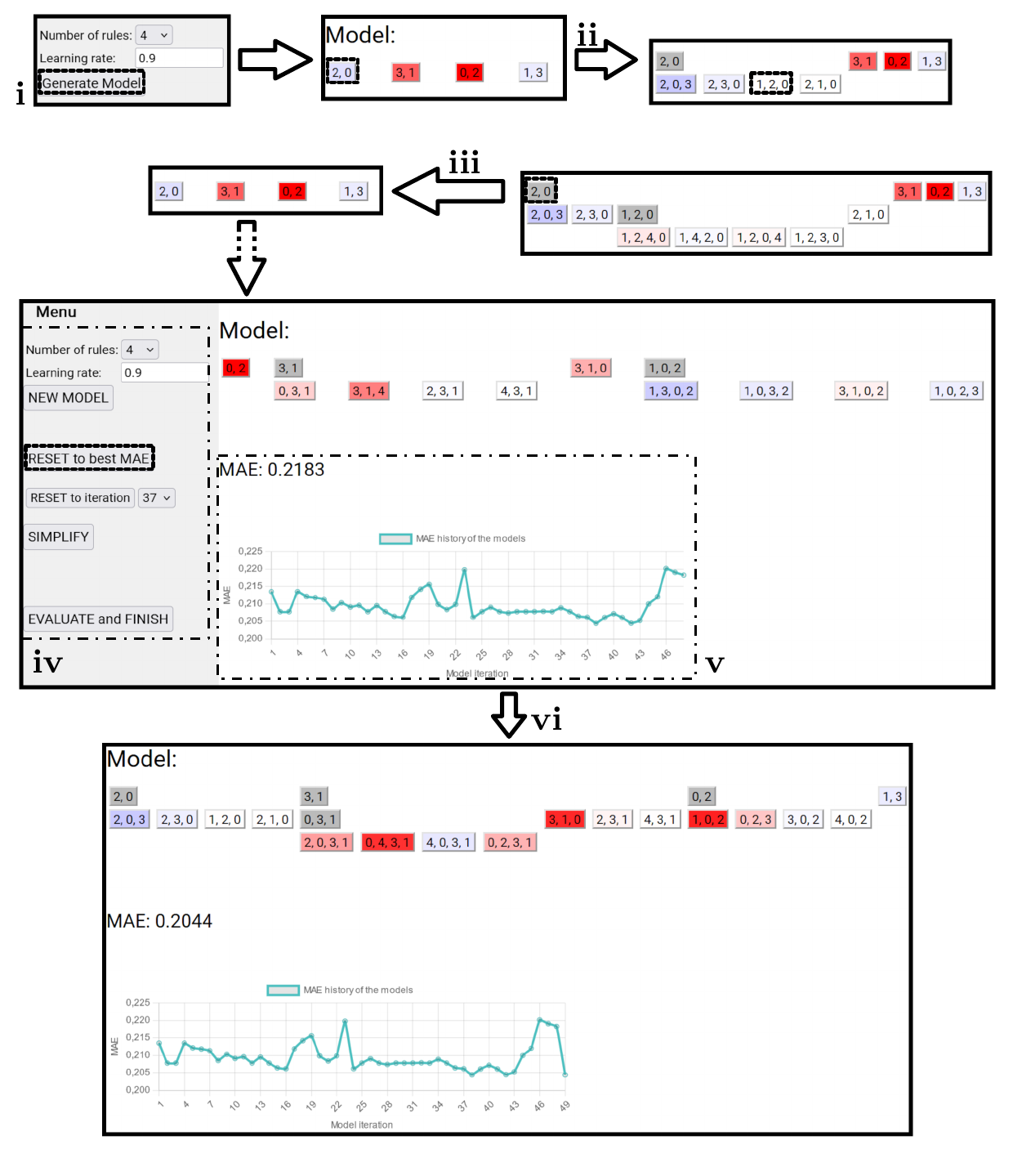}}
    \caption{Exemplary use of HuGuR. The first step \textbf{i} is to select the hyperparameters, which are the number of constraints added in each step and the learning rate of the model. The number of constraints $l$ is an integer between 1 and 20 and the learning rate is a float between 0.000001 and 1. After the generation, we obtain the model view. Here, we see the constraints the model encompasses, and we can interact with them. Blue constraints indicate a positive impact on the target value and red constraints indicate a negative impact. By clicking on an active constraint \textbf{ii}, we deactivate the constraint and replace it by the $l$ most promising child constraints. And vice versa, by clicking on an inactive constraint \textbf{iii}, we activate it and remove all of its child constraints. We can also use additional controls \textbf{iv} to restart with other hyperparameters or reset to a previous iteration. The error history \textbf{v} lets us track our progress over the iterations. Also, one can always jump back to the best model found so far \textbf{vi}.}
    \label{fig:interface}
\end{figure}

To further increase the comprehensibility of the model from Section \ref{transpermreg} and enable potential users to be able to benefit from their own background knowledge, we want to incorporate their input in the creation process of the model\footnote{An interactive demo can be found on: hugur.pensel.eu}. Figure \ref{fig:interface} shows multiple screenshots of the HuGuR interface and exemplifies its use.

\begin{algorithm}[t!]
\caption{Select $l$ constraints}\label{alg:selconstraints}
\KwData{Permutations $X$, set of constraints $\rho$, residuals $\delta$, number of selected constraints $l$}
\KwResult{most promising constraints $\rho^*$}
$\rho^* \gets \{\}$\;
$\rho' \gets \{\}$\;
$T \gets \{\}$\;
\For{$\varrho \in \rho$}{
    $Z \gets g_{\varrho}(X)$\;
    $\tau \gets |\sum \delta Z|$\;
    $\rho'.append(\varrho)$\;
    $T.append(\tau)$\;
}
$I \gets argsort(T)$\Comment*[r]{$argsort(T)_i:=\{j|(\tau_i < \tau_j) \text{ or } (\tau_i = \tau_j, i<j) \}$}
\For{$i \in \{I_{|I|-l},\dots,I_{|I|}\}$}{
$\rho^*.append(\varrho_i \in \rho')$
}
\end{algorithm}

The HIL model HuGuR initially generates $l$ minimal constraints, which are constraints consisting of only two elements, by using the set of all possible minimal constraints as input for Algorithm \ref{alg:selconstraints}. We use the average target value $\mu$ of the provided training set as base prediction to obtain the first residuals and calculate the coefficients $\beta$ for the constraints one after another by the gradient boosting method introduced in Section \ref{transpermreg}. Then, we present the selected constraints and their corresponding coefficients, as well as the current prediction error on a validation set, to the user. To improve the comprehensibility, we provide the coefficients by the hue and saturation of the color for each constraint. The saturation corresponds to the absolute value of the coefficient and the hue indicates a positive or negative impact on the predicted target, with blue corresponding to positive and red to negative. For instance, if we predict the difference from an optimal permutation, negative coefficients are blue and positive red, since a smaller difference is better.

\begin{algorithm}[t!]
\caption{Generate all child constraints}\label{alg:genchild}
\KwData{Item set $S$, prior constraint $\varrho'$}
\KwResult{All child constraints $\rho'$}
$\rho' \gets []$\;
\For{$c \in S$}{
\If{$c \notin \varrho'$}{
\For{$i \in \{1,\dots,|\varrho'|+1\}$}{
$\varrho^+ \gets (\varrho'_1,\dots,\varrho'_{i-1},c,\varrho'_i,\dots,\varrho'_{|\varrho'|})$\;
$\rho'[] \gets \varrho^+$\;
}
}
}
\end{algorithm}

The user can further refine the model by replacing a single constraint $\varrho'$ with $l$ more specific child constraints. First, we generate all possible child constraints $\rho'$ as in Algorithm \ref{alg:genchild}, and then we select the most promising $l$ constraints of those by Algorithm \ref{alg:selconstraints}. To calculate the coefficients,  we remove $\varrho$ from the constraints of the model $\rho$ and use the altered prediction to obtain the residuals. Alternatively, the user can generate a new model with different hyperparameters or simplify the current model. To simplify a model, we choose the $l$ constraints with the highest absolute coefficients and use them as the foundation of a new model. Additionally, the user can reverse this by adding $\varrho$ back to $\rho$ and removing all of its children. During the whole process we track the mean absolute error on the validation set and present the histories to the user. The user can always reload any previous model, in particular the model with the lowest error.

\section{Data}
\label{data}

Each data set used in this paper consists of a set of permutations $X$ and a corresponding set of numerical values $Y$, where $x_i \in X$ relates to $y_i \in Y$. Some statistics of the different data sets are shown in Table \ref{tab:datasets}. We list the number of permutations in the data set as size together with the sizes of the train, validation and test splits, the number of items in each underlying set as features, the number of unique permutations in the data set, and the mean and standard deviation of the target variables.

\begin{table}
\begin{center}
\caption{Statistics of the different data sets.}
\label{tab:datasets}
\resizebox{\linewidth}{!}{\begin{tabular}{l|ccc|cc}
\toprule
Data set & Size (train/validation/test) & Features & Unique permutations & Mean($y$) & Std($y$) \\
\midrule
$edm_5\_small$ & 750(450/50/250) & 5 & 74 & 0.716 & 0.282 \\
$edm_9\_small$ & 750(450/50/250) & 9 & 332 & 0.665 & 0.258 \\
$edm_{18}\_small$ & 750(450/50/250) & 18 & 630 & 0.651 & 0.249 \\
$cc_{enron}$ & 3000(1800/200/1000) & 53 & 3000 & 0.427 & 0.005 \\
$cc_{yeast}\_small$ & 7500(4500/500/2500) & 14 & 7500 & 0.484 & 0.011 \\
\hline
$edm_5\_large$ & 2928(1728/200/1000) & 5 & 116 & 0.718 & 0.280 \\
$edm_9\_large$ & 2907(1707/200/1000) & 9 & 962 & 0.667 & 0.258 \\
$edm_{18}\_large$ & 2644(1444/200/1000) & 18 & 2095 & 0.644 & 0.255 \\
$cc_{yeast}\_large$ & 15000(9000/1000/5000) & 14 & 15000 & 0.484 & 0.011 \\
        \bottomrule
\end{tabular}}
\end{center}

\end{table}

\subsection{Classifier Chain Orderings}

As the first real-world problem for permutation regression, we chose the ordering of classifer chains \cite{classifierchains}. Classifier chains is a problem transformation method for multi-label classification. For a set of $n$ labels $L = \{l_1, \dots, l_n \}$ a chain of $n$ classifiers $CC = \{C_1, \dots, C_n \}$ are trained, where $C_i$ predicts $l_i$ given the instance features $x$ and the previously predicted labels $l_{<i}$. Since each label has to be predicted exactly once by the classifier chain, the ordering of the chain is a permutation of the classifiers. The classification performance of a specific ordering can be measured and therefore be used as regression target.

The first data set used to generate classifier chain orderings is the yeast  \cite{yeast} multi-label data set, which has 14 different labels. Additionally we generated classifier chain orderings for the enron \cite{enron} multi-label data set, which has 53 different labels. Both data sets were downloaded from the OpenML \cite{openml} data repository.

The classifier chains are trained and evaluated using scikit-learn \cite{scikit-learn} implementations. For the classifer chain, we used \textit{sklearn.multioutput.ClassifierChain} with \textit{sklearn.linear\_model.LogisticRegression} as their elements. The performance of the chains is measured by the \textit{sklearn.metrics.jaccard\_score}, which calculates the Jaccard index \cite{jaccard} of the true and the predicted label set. For each data set we randomly generated 100000 classifier chain orderings and computed the corresponding scores as target values.

In order to be able to run the data sets with our online survey setup, we only use a small randomly sampled fraction of the data sets in this work. We sampled 3000 instances for $CC_{enron}$, 7500 instances for $cc_{yeast}\_small$ and 15000 instances for $cc_{yeast}\_large$.

\subsection{Educational Domain}

The second real-world area of application for permutation regression we chose is learning analytics. The ordering of different courses in a study, the order of different tasks in a course, or the order of different questions in a test are permutations of an item set and the performance of a student can be used as corresponding target value.

As base for this data sets we used the NeurIPS 2020 Education Challenge data set for task 3 and 4 \cite{neurips_data}. Using FP-growth \cite{fpgrowth}, we select three differently sized sets of questions with a sufficiently large minimum support as the base item sets for the permutations. For each user that answered all selected question, we order the questions by the answer timestamp and use the fraction of correctly answered questions as target value.

This yields the $edm_5$, $edm_9$ and $edm_{18}$ data sets, which are each built on a set of five, nine and 18 questions. The large versions of those data sets comprise all available instances of those specific sets of questions, and for the small versions, we use a random sampling of 750 instances.
\section{Experiments}
\label{experiments}

To measure the performance of HuGuR in its designated field of application, we conduct a user study with 21 participants. The participants consist of students, mostly computer science, with a varying degree of experience in the field of machine learning.

\subsection{User Study Setup}
We split the nine data sets presented in Section \ref{data} each in single independent train, validation and test sets. Each participant is assigned to work on a random and not previously seen data set. The participants receive no further information about the data sets, apart from the information conveyed by the HuGuR interface. In this way, we leveled
the playing field for the experiment for all participants.
After each finalization of their model, the participant can choose to continue with a new random data set, until all nine data sets are finished. We encourage the participants to finish at least four data sets and to go through at least 20 model iterations during the training phase. Therefore, we only consider models with a minimum exploration of at least 20 model iteration steps in our presented results. This yields at least 12 different models for each of the nine data sets. To ensure a performance-oriented participation, we gave a monetary bonus to those participants who produced the highest performing models for each data set. We assume that users have a similar incentive to build good models when they construct models for commercial or research purposes.

\subsection{User Study Results}

\begin{table}[]
    \centering
    \caption{Hyperparameter grid for the neural network based comparison methods. The $layers$ parameter describes the structure of the dense prediction part of the networks. The number of units in the sequence encoding LSTM part of the model is given by $encoder\_units$ parameter and the $alpha$ parameter sets the strength of the L2 regularization term.}
    \label{tab:hyperparams}
    \begin{tabular}{c|ll}
    \toprule
        Algorithm & Parameter & Values \\
        \midrule
        $ENC_{one-hot}$ & $layers$ & (50,), (100,), (100,50), (100,100,50), (200,100,50) \\
        $ENC_{graph}$ & $alpha$ & 0.001, 0.0001, 0.00001 \\
        \hline
        \multirow{ 2}{*}{$LSTM$} & $layers$ & (50,), (100,), (100,50), (100,100,50), (200,100,50) \\
        & $encoder\_units$ & 10, 25, 50, 100\\
        \bottomrule
    \end{tabular}

\end{table}

We calculate the mean absolute error, mean squared error, and the coefficient of determination ($R^2$ score) for all user-built HuGuR models. We also train multiple models with our comparison methods on the train splits of our data sets. For the simple one-hot and graph encoding based approaches, we employ the scikit-learn \cite{scikit-learn} multilayer perceptron (MLP) as regressor, and for the sequence based encoding,  we build a more complex neural network regressor using tensorflow \cite{tensorflow} and keras. We also incorporate a small hyperparameter search over the parameter grid provided in Table \ref{tab:hyperparams}, and evaluate the solution candidates on a dedicated validation set. The comparison models are similarly evaluated on the test splits of the data sets\footnote{The source code and the data sets are available on github.com/lpensel/HuGuR}. 

In addition to the average performance of all HuGuR models on the data sets, we also compare the average performance of the five HuGuR models ($HuGuR_5$), i.e., the models of the best five users, which yield the best performance on the validation sets to all the other methods.

First, we compare the performance of HuGuR with the naive baseline and the two simpler encoding methods. Since we have multiple HuGuR models for each data set, we provide the average performance as well as its standard deviation as result. The performance measures are compiled in Table \ref{tab:mae1}. Considering the smaller data sets, which are always ordered to the top of the tables, HuGuR outperforms the comparison methods on almost all of those data sets. With an average rank over all metrics of 1.4 for $HuGuR_5$ and 2.13 for $HuGuR$ compared to a rank of 3.53 for $ENC_{graph}$ and 3.73 for $ENC_{one-hot}$. Adding the larger data sets back into the consideration, the average ranks change to 1.7 and 2.67 for $HuGuR_5$ and $HuGuR$ as well as to 2.96 for $ENC_{one-hot}$ and 3.11 for $ENC_{graph}$. Using an adapted Friedman test \cite{ranking:signific}, it shows that there are significant ($p << 0.001$) differences in the distribution of ranks, i.e. the quality of the methods. To determine significant differences between the algorithms, we use the Bonferroni-Dunn test \cite{ranking:signific} with a significance threshold of $\alpha = 0.05$ and present the results in Table \ref{tab:rank_difference} under $Set_1$.

\begin{table}[]
    \centering
    \caption{Performance measures for the naive baseline $NB$, one-hot encoding with MLP regressor $ENC_{one-hot}$, graph encoding with MLP regressor $ENC_{graph}$, all user-built models $HuGuR$ and the top five user-built models $HuGuR_5$. The best performing method for each data set is highlighted. Additionally, we provide the average rank for each method based on these results.}
    \label{tab:mae1}
    \begin{tabular}{l|ccccc}
    \toprule
    \multicolumn{6}{c}{\textbf{Mean absolute errors} } \\
    \midrule
          Data set & $NB$ & $ENC_{one-hot}$ & $ENC_{graph}$ & $HuGuR$ & $HuGuR_5$ \\ 
\midrule 
$edm_5\_small$ & $2.506e^{-1} $& $\mathbf{ 2.394e^{-1} } $& $2.468e^{-1} $& $2.409e^{-1}(4.422e^{-3}) $& $2.395e^{-1}(4.422e^{-3}) $\\ 
$edm_9\_small$ & $2.197e^{-1} $& $2.163e^{-1} $& $2.183e^{-1} $& $2.110e^{-1}(1.953e^{-3}) $& $\mathbf{ 2.109e^{-1}(1.953e^{-3}) } $\\ 
$edm_{18}\_small$ & $2.086e^{-1} $& $2.001e^{-1} $& $\mathbf{ 1.686e^{-1} } $& $1.815e^{-1}(8.729e^{-3}) $& $1.758e^{-1}(8.729e^{-3}) $\\ 
$cc_{enron}$ & $4.270e^{-3} $& $7.384e^{-3} $& $4.720e^{-3} $& $3.468e^{-3}(1.204e^{-4}) $& $\mathbf{ 3.426e^{-3}(1.204e^{-4}) } $\\ 
$cc_{yeast}\_small$ & $9.105e^{-3} $& $7.563e^{-3} $& $8.749e^{-3} $& $6.948e^{-3}(6.816e^{-4}) $& $\mathbf{ 6.315e^{-3}(6.816e^{-4}) } $\\ 
\hline
$edm_5\_large$ & $2.368e^{-1} $& $\mathbf{ 2.144e^{-1} } $& $2.180e^{-1} $& $2.205e^{-1}(7.126e^{-3}) $& $2.145e^{-1}(7.126e^{-3}) $\\ 
$edm_9\_large$ & $2.116e^{-1} $& $2.013e^{-1} $& $\mathbf{ 1.964e^{-1} } $& $1.992e^{-1}(5.124e^{-3}) $& $1.967e^{-1}(5.124e^{-3}) $\\ 
$edm_{18}\_large$ & $2.173e^{-1} $& $1.753e^{-1} $& $\mathbf{ 1.734e^{-1} } $& $1.873e^{-1}(1.054e^{-2}) $& $1.812e^{-1}(1.054e^{-2}) $\\ 
$cc_{yeast}\_large$ & $8.903e^{-3} $& $\mathbf{ 5.458e^{-3} } $& $8.425e^{-3} $& $6.901e^{-3}(7.311e^{-4}) $& $6.158e^{-3}(7.311e^{-4}) $\\ 
\hline \hline 
 Average rank & 4.78 & 2.67 & 2.89 & 2.89 & 1.78 \\

    \toprule
    \multicolumn{6}{c}{\textbf{Mean squared errors} } \\
    \midrule
          Data set & $NB$ & $ENC_{one-hot}$ & $ENC_{graph}$ & $HuGuR$ & $HuGuR_5$ \\ 
 \midrule
$edm_5\_small$ & $8.921e^{-2} $& $9.015e^{-2} $& $9.041e^{-2} $& $8.610e^{-2}(2.533e^{-3}) $& $\mathbf{ 8.606e^{-2}(2.533e^{-3}) } $\\ 
$edm_9\_small$ & $6.915e^{-2} $& $6.653e^{-2} $& $6.741e^{-2} $& $\mathbf{ 6.246e^{-2}(1.279e^{-3}) } $& $6.289e^{-2}(1.279e^{-3}) $\\ 
$edm_{18}\_small$ & $5.804e^{-2} $& $6.466e^{-2} $& $\mathbf{ 4.577e^{-2} } $& $4.968e^{-2}(2.734e^{-3}) $& $4.853e^{-2}(2.734e^{-3}) $\\ 
$cc_{enron}$ & $2.842e^{-5} $& $8.736e^{-5} $& $3.612e^{-5} $& $1.912e^{-5}(1.413e^{-6}) $& $\mathbf{ 1.868e^{-5}(1.413e^{-6}) } $\\ 
$cc_{yeast}\_small$ & $1.283e^{-4} $& $9.012e^{-5} $& $1.174e^{-4} $& $7.705e^{-5}(1.463e^{-5}) $& $\mathbf{ 6.363e^{-5}(1.463e^{-5}) } $\\ 
\hline
$edm_5\_large$ & $8.007e^{-2} $& $\mathbf{ 7.321e^{-2} } $& $7.522e^{-2} $& $7.448e^{-2}(1.951e^{-3}) $& $7.344e^{-2}(1.951e^{-3}) $\\ 
$edm_9\_large$ & $6.610e^{-2} $& $6.321e^{-2} $& $5.825e^{-2} $& $5.852e^{-2}(2.906e^{-3}) $& $\mathbf{ 5.745e^{-2}(2.906e^{-3}) } $\\ 
$edm_{18}\_large$ & $6.398e^{-2} $& $4.724e^{-2} $& $\mathbf{ 4.485e^{-2} } $& $5.167e^{-2}(4.357e^{-3}) $& $4.953e^{-2}(4.357e^{-3}) $\\ 
$cc_{yeast}\_large$ & $1.242e^{-4} $& $\mathbf{ 4.770e^{-5} } $& $1.109e^{-4} $& $7.653e^{-5}(1.559e^{-5}) $& $6.034e^{-5}(1.559e^{-5}) $\\  
\hline \hline 
 Average rank & 4.44 & 3.11 & 3.22 & 2.56 & 1.67\\

    \toprule
    \multicolumn{6}{c}{\textbf{$R^2$ scores} } \\
    \midrule
         Data set & $NB$ & $ENC_{one-hot}$ & $ENC_{graph}$ & $HuGuR$ & $HuGuR_5$ \\ 
\midrule 
$edm_5\_small$ & $-1.282e^{-2} $& $-2.355e^{-2} $& $-2.647e^{-2} $& $2.246e^{-2}(2.876e^{-2}) $& $\mathbf{ 2.285e^{-2}(2.876e^{-2}) } $\\ 
$edm_9\_small$ & $-8.047e^{-4} $& $3.713e^{-2} $& $2.441e^{-2} $& $\mathbf{ 9.608e^{-2}(1.852e^{-2}) } $& $8.979e^{-2}(1.852e^{-2}) $\\ 
$edm_{18}\_small$ & $-1.283e^{-2} $& $-1.283e^{-1} $& $\mathbf{ 2.013e^{-1} } $& $1.331e^{-1}(4.770e^{-2}) $& $1.531e^{-1}(4.770e^{-2}) $\\ 
$cc_{enron}$ & $-8.550e^{-4} $& $-2.076e+00 $& $-2.720e^{-1} $& $3.266e^{-1}(4.976e^{-2}) $& $\mathbf{ 3.420e^{-1}(4.976e^{-2}) } $\\ 
$cc_{yeast}\_small$ & $-1.538e^{-4} $& $2.977e^{-1} $& $8.509e^{-2} $& $3.995e^{-1}(1.140e^{-1}) $& $\mathbf{ 5.041e^{-1}(1.140e^{-1}) } $\\ 
\hline
$edm_5\_large$ & $-3.862e^{-4} $& $\mathbf{ 8.533e^{-2} } $& $6.029e^{-2} $& $6.947e^{-2}(2.437e^{-2}) $& $8.253e^{-2}(2.437e^{-2}) $\\ 
$edm_9\_large$ & $-4.697e^{-5} $& $4.365e^{-2} $& $1.187e^{-1} $& $1.146e^{-1}(4.396e^{-2}) $& $\mathbf{ 1.309e^{-1}(4.396e^{-2}) } $\\ 
$edm_{18}\_large$ & $-1.639e^{-5} $& $2.617e^{-1} $& $\mathbf{ 2.990e^{-1} } $& $1.924e^{-1}(6.810e^{-2}) $& $2.258e^{-1}(6.810e^{-2}) $\\ 
$cc_{yeast}\_large$ & $-8.051e^{-5} $& $\mathbf{ 6.160e^{-1} } $& $1.074e^{-1} $& $3.840e^{-1}(1.255e^{-1}) $& $5.143e^{-1}(1.255e^{-1}) $\\ 
\hline \hline 
 Average rank & 4.44 & 3.11 & 3.22 & 2.56 & 1.67   \\

 \hline \hline
 Overall rank & 4.56 & 2.96 & 3.11 & 2.67 & 1.7 \\
        \bottomrule
    \end{tabular}

\end{table}

The comparisons of HuGuR with the more complex sequence based encoding method are gathered in Table \ref{tab:mae2}. While $HuGuR_5$ slightly outperforms $LSTM$ on the small data sets, rank 1.7 opposed to rank 1.9, on all data sets combined, $LSTM$ achieves an average rank of 1.67 and $HuGuR_5$ only an average rank of 1.74. The adapted Friedman test still shows a significant ($p < 0.001$) difference between the rankings of $LSTM$, $HuGuR$ and $HuGuR_5$. Table \ref{tab:rank_difference} lists the significant differences under $Set_2$, based on a Bonferroni-Dunn test with a threshold of $\alpha = 0.05$.

\begin{table}[]
    \centering
    \caption{Performance measures for the sequence encoding with LSTM based neural network architecture $LSTM$, all user-built models $HuGuR$ and the top five user build models $HuGuR_5$. The best performing method for each data set is highlighted. Additionally, we provide the average rank for each method based on these results.}
    \label{tab:mae2}
    \begin{tabular}{l|ccc}
    \toprule
    \multicolumn{4}{c}{\textbf{Mean absolute errors} } \\
    \midrule
         Data set & $LSTM$ & $HuGuR$ & $HuGuR_5$ \\ 
\midrule
$edm_5\_small$ & $\mathbf{ 2.364e^{-1} } $& $2.409e^{-1}(4.422e^{-3}) $& $2.395e^{-1}(4.422e^{-3}) $\\ 
$edm_9\_small$ & $\mathbf{ 2.005e^{-1} } $& $2.110e^{-1}(1.953e^{-3}) $& $2.109e^{-1}(1.953e^{-3}) $\\ 
$edm_{18}\_small$ & $\mathbf{ 1.709e^{-1} } $& $1.815e^{-1}(8.729e^{-3}) $& $1.758e^{-1}(8.729e^{-3}) $\\ 
$cc_{enron}$ & $4.251e^{-3} $& $3.468e^{-3}(1.204e^{-4}) $& $\mathbf{ 3.426e^{-3}(1.204e^{-4}) } $\\ 
$cc_{yeast}\_small$ & $\mathbf{ 4.671e^{-3} } $& $6.948e^{-3}(6.816e^{-4}) $& $6.315e^{-3}(6.816e^{-4}) $\\ 
\hline
$edm_5\_large$ & $\mathbf{ 2.084e^{-1} } $& $2.205e^{-1}(7.126e^{-3}) $& $2.145e^{-1}(7.126e^{-3}) $\\ 
$edm_9\_large$ & $\mathbf{ 1.823e^{-1} } $& $1.992e^{-1}(5.124e^{-3}) $& $1.967e^{-1}(5.124e^{-3}) $\\ 
$edm_{18}\_large$ & $\mathbf{ 1.594e^{-1} } $& $1.873e^{-1}(1.054e^{-2}) $& $1.812e^{-1}(1.054e^{-2}) $\\ 
$cc_{yeast}\_large$ & $\mathbf{ 3.737e^{-3} } $& $6.901e^{-3}(7.311e^{-4}) $& $6.158e^{-3}(7.311e^{-4}) $\\ 
\hline \hline 
 Average rank & 1.22 & 2.89 & 1.89   \\
        
    \toprule
    \multicolumn{4}{c}{\textbf{Mean squared errors} } \\
    \midrule
          Data set & $LSTM$ & $HuGuR$ & $HuGuR_5$ \\ 
\midrule
$edm_5\_small$ & $8.904e^{-2} $& $8.610e^{-2}(2.533e^{-3}) $& $\mathbf{ 8.606e^{-2}(2.533e^{-3}) } $\\ 
$edm_9\_small$ & $\mathbf{ 5.838e^{-2} } $& $6.246e^{-2}(1.279e^{-3}) $& $6.289e^{-2}(1.279e^{-3}) $\\ 
$edm_{18}\_small$ & $5.090e^{-2} $& $4.968e^{-2}(2.734e^{-3}) $& $\mathbf{ 4.853e^{-2}(2.734e^{-3}) } $\\ 
$cc_{enron}$ & $2.841e^{-5} $& $1.912e^{-5}(1.413e^{-6}) $& $\mathbf{ 1.868e^{-5}(1.413e^{-6}) } $\\ 
$cc_{yeast}\_small$ & $\mathbf{ 3.391e^{-5} } $& $7.705e^{-5}(1.463e^{-5}) $& $6.363e^{-5}(1.463e^{-5}) $\\ 
\hline
$edm_5\_large$ & $7.590e^{-2} $& $7.448e^{-2}(1.951e^{-3}) $& $\mathbf{ 7.344e^{-2}(1.951e^{-3}) } $\\ 
$edm_9\_large$ & $\mathbf{ 5.314e^{-2} } $& $5.852e^{-2}(2.906e^{-3}) $& $5.745e^{-2}(2.906e^{-3}) $\\ 
$edm_{18}\_large$ & $\mathbf{ 4.194e^{-2} } $& $5.167e^{-2}(4.357e^{-3}) $& $4.953e^{-2}(4.357e^{-3}) $\\ 
$cc_{yeast}\_large$ & $\mathbf{ 2.223e^{-5} } $& $7.653e^{-5}(1.559e^{-5}) $& $6.034e^{-5}(1.559e^{-5}) $\\ 
\hline \hline 
 Average rank & 1.89 & 2.44 & 1.67 \\
        
    \toprule
    \multicolumn{4}{c}{\textbf{$R^2$ scores} } \\
    \midrule
         Data set & $LSTM$ & $HuGuR$ & $HuGuR_5$ \\ 
\midrule
$edm_5\_small$ & $-1.098e^{-2} $& $2.246e^{-2}(2.876e^{-2}) $& $\mathbf{ 2.285e^{-2}(2.876e^{-2}) } $\\ 
$edm_9\_small$ & $\mathbf{ 1.551e^{-1} } $& $9.608e^{-2}(1.852e^{-2}) $& $8.979e^{-2}(1.852e^{-2}) $\\ 
$edm_{18}\_small$ & $1.119e^{-1} $& $1.331e^{-1}(4.770e^{-2}) $& $\mathbf{ 1.531e^{-1}(4.770e^{-2}) } $\\ 
$cc_{enron}$ & $-4.006e^{-4} $& $3.266e^{-1}(4.976e^{-2}) $& $\mathbf{ 3.420e^{-1}(4.976e^{-2}) } $\\ 
$cc_{yeast}\_small$ & $\mathbf{ 7.357e^{-1} } $& $3.995e^{-1}(1.140e^{-1}) $& $5.041e^{-1}(1.140e^{-1}) $\\ 
\hline
$edm_5\_large$ & $5.183e^{-2} $& $6.947e^{-2}(2.437e^{-2}) $& $\mathbf{ 8.253e^{-2}(2.437e^{-2}) } $\\ 
$edm_9\_large$ & $\mathbf{ 1.960e^{-1} } $& $1.146e^{-1}(4.396e^{-2}) $& $1.309e^{-1}(4.396e^{-2}) $\\ 
$edm_{18}\_large$ & $\mathbf{ 3.444e^{-1} } $& $1.924e^{-1}(6.810e^{-2}) $& $2.258e^{-1}(6.810e^{-2}) $\\ 
$cc_{yeast}\_large$ & $\mathbf{ 8.211e^{-1} } $& $3.840e^{-1}(1.255e^{-1}) $& $5.143e^{-1}(1.255e^{-1}) $\\ 
\hline \hline 
 Average rank & 1.89 & 2.44 & 1.67  \\

 \hline \hline
 Overall rank & 1.67 & 2.59 & 1.74\\
        \bottomrule
    \end{tabular}

\end{table}

Comparing all methods with each other gives again significant ($p << 0.001$) differences in the rank distributions and using the Bonferroni-Dunn test with a threshold $\alpha = 0.05$. Table \ref{tab:rank_difference} under $Set_3$ shows that $LSTM$ performs significantly better than all other methods except $HuGuR_5$. Similarly, $HuGuR_5$ performs significantly better than all other methods but $LSMT$ and $HuGuR$.

Overall, HuGuR achieves a similar or better performance than the baseline methods, while not being a black-box neural network based model but a linear model with human-understandable constraints. Additionally, if we compare the complexity of the trained models, all HuGuR models considered in our study have at least one order of magnitude fewer trainable parameters than the neural network based methods for the same data set. The biggest HuGuR model comprises 460 coefficients, while the smallest $LSTM$ model comprises 1401 weights and biases.

Table \ref{tab:model_comp} compares the number of trainable parameters of all models. In all cases HuGuR has at least one order of magnitude fewer trainable parameters than the other methods and often even multiple orders of magnitude fewer parameters.

\begin{table}[t!]
    \centering
    \caption{Statistical significant differences of the algorithms based on their ranks for the performance measures. We use the Bonferroni-Dunn test with $\alpha = 0.05$ to show the significance. $Set_1$ compares $NB$, $ENC_{one-hot}$, $ENC_{graph}$, $HuGuR$ and $HuGuR_5$. $Set_2$ compares $LSTM$, $HuGuR$ and $HuGuR_5$. $Set_3$ compares all methods.
    }
    \label{tab:rank_difference}
    \begin{tabular}{rcl|rcl|rcl}
    \toprule
       \multicolumn{3}{c|}{$Set_1$}  & \multicolumn{3}{c|}{$Set_2$} & \multicolumn{3}{c}{$Set_3$} \\
       \midrule
         $ENC_{one-hot}$ &  $>$  & $NB$ & $LSTM$ &  $>$  & $HuGuR$  & \hspace{5pt} $ENC_{one-hot}$ &  $>$  & $NB$ \\
$ENC_{graph}$ & $>$ & $NB$ & \hspace{5pt} $HuGuR_5$ & $>$ & $HuGuR$ \hspace{5pt} & $ENC_{graph}$ &  $>$  & $NB$ \\
$HuGuR$ & $>$ & $NB$ & & &  & $LSTM$ &  $>$  & $NB$ \\
$HuGuR_5$ & $>$ & $NB$ & & & & $HuGuR$ &  $>$  & $NB$ \\
$HuGuR_5$ & $>$ & $ENC_{one-hot}$ \hspace{5pt} & & & & $HuGuR_5$ &  $>$  & $NB$ \\
$HuGuR_5$ & $>$ & $ENC_{graph}$ & & & & $LSTM$ &  $>$  & $ENC_{one-hot}$ \\
& & & & & & $HuGuR_5$ &  $>$  & $ENC_{one-hot}$ \\
& & & & & & $LSTM$ &  $>$  & $ENC_{graph}$ \\
& & & & & & $HuGuR_5$ &  $>$  & $ENC_{graph}$ \\
& & & & & & $LSTM$ &  $>$  & $HuGuR$\\
        \bottomrule
    \end{tabular}

\end{table}

\begin{table}[]
    \centering
    \caption{Number of trainable parameters, i.e. the model complexity, for each model compared in our experiments. For $HuGuR$ and $HuGuR_5$ we provide the average and standard deviation for the number of parameters of all used models.}
    \label{tab:model_comp}
    \begin{tabular}{l|cccccc}
    \toprule
         Data set  & $ENC_{one-hot}$ & $ENC_{graph}$ & $LSTM$ & $HuGuR$ & $HuGuR_5$  \\
         \midrule
$edm_5\_small$   &  2701  &  1851  &  1241  & 90.25(54.49) & 112.40(22.84) \\ 
$edm_9\_small$   &  41601  &  45201  &  17101  & 65.75(48.72) & 47.60(16.32) \\ 
$edm_{18}\_small$   &  37601  &  51301  &  2361  & 147.14(171.09) & 234.80(236.92) \\ 
$cc_{enron}$ &   296201  &  608401  &  23401  & 98.00(49.46) & 67.20(24.03) \\ 
$cc_{yeast}\_small$   &  64601  &  70201  &  56201  & 171.83(114.31) & 246.40(112.02) \\ 
\hline 
$edm_5\_large$   &  17801  &  1851  &  57601  & 64.62(40.10) & 92.60(37.40) \\ 
$edm_9\_large$   &  4151  &  15101  &  1401  & 108.42(55.39) & 152.00(39.70) \\ 
$edm_{18}\_large$   &  90201  &  97401  &  2361  & 146.77(157.55) & 222.20(221.48) \\ 
$cc_{yeast}\_large$   &  64601  &  70201  &  51101  & 162.17(138.96) &  262.40(140.57)\\
        \bottomrule
    \end{tabular}

\end{table}

\subsection{User Study Evaluation}

\begin{figure}[!htbp]
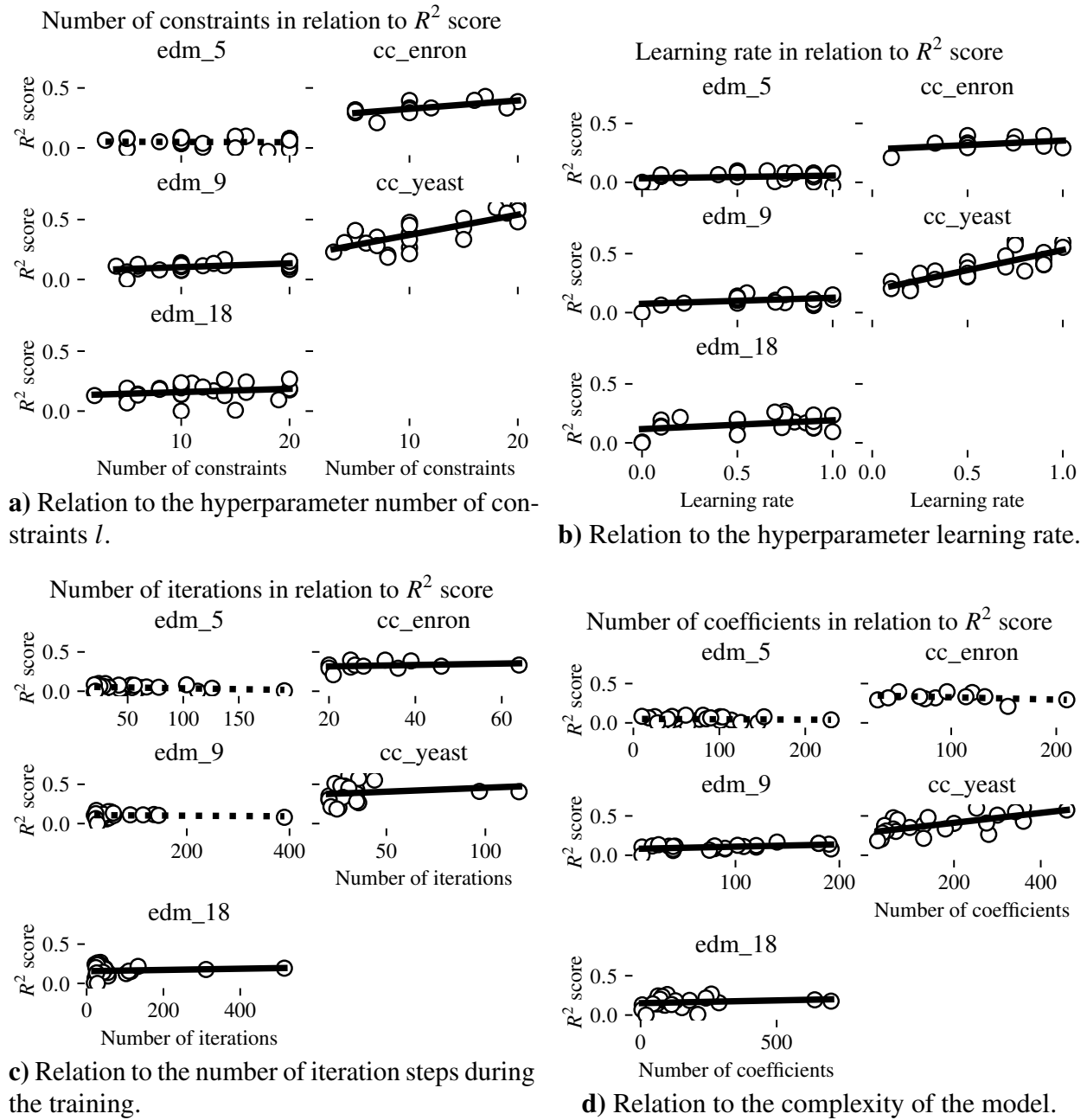

    \centering
    \begin{subfigure}{0.49\textwidth}
        \resizebox{\linewidth}{!}{\input{nr_to_r2.pgf}}
       \caption{Relation to the hyperparameter number of constraints $l$.}
        \label{fig:nr_to_r2}
    \end{subfigure}
    \hfill
  \begin{subfigure}{0.49\textwidth}
        \resizebox{\linewidth}{!}{\input{lr_to_r2.pgf}}
        \caption{Relation to the hyperparameter learning rate.}
        \label{fig:lr_to_r2}
    \end{subfigure}
    \hfill
    \begin{subfigure}{0.49\textwidth}
        \resizebox{\linewidth}{!}{\input{ni_to_r2.pgf}}
        \caption{Relation to the number of iteration steps during the training.}
        \label{fig:ni_to_r2}
    \end{subfigure}
    \hfill
   \begin{subfigure}{0.49\textwidth}
        \resizebox{\linewidth}{!}{\input{np_to_r2.pgf}}
        \caption{Relation to the complexity of the model.}
        \label{fig:np_to_r2}
    \end{subfigure}

    \caption{For each data set -- data sets with small and large versions such as $edm_5$ are combined into one plot -- we examine the relation between the measured performance, here the $R^2$ score, and multiple properties of the user-built models. Each point represents one human-built model. A dotted line represents a negative slope of the line of best fit, while a solid line represents a positive slope.}
    \label{fig:vals_to_r2}
\end{figure}

In addition to comparing the performance of the human-built models with the baseline models, we also evaluated some potential influences of different model properties on the performance of a model. In Figure \ref{fig:vals_to_r2}, we scatter the $R^2$ score of a model against one of its properties and plot the line of best linear fit. Figure \ref{fig:lr_to_r2} indicates that overall a higher learning rate leads to a better performance, and Figure \ref{fig:nr_to_r2} shows that for the most data sets a higher number of constraints $l$ added in each step increases the performance. In Figure \ref{fig:ni_to_r2}, the relation of the $R^2$ score to the number of iterations of refinement by a user is presented and Figure \ref{fig:np_to_r2} depicts the relation between performance and the number of coefficients in the model (i.e. the complexity of the model).

\begin{figure}[!ht]
    \centering
    \resizebox{\linewidth}{!}{\input{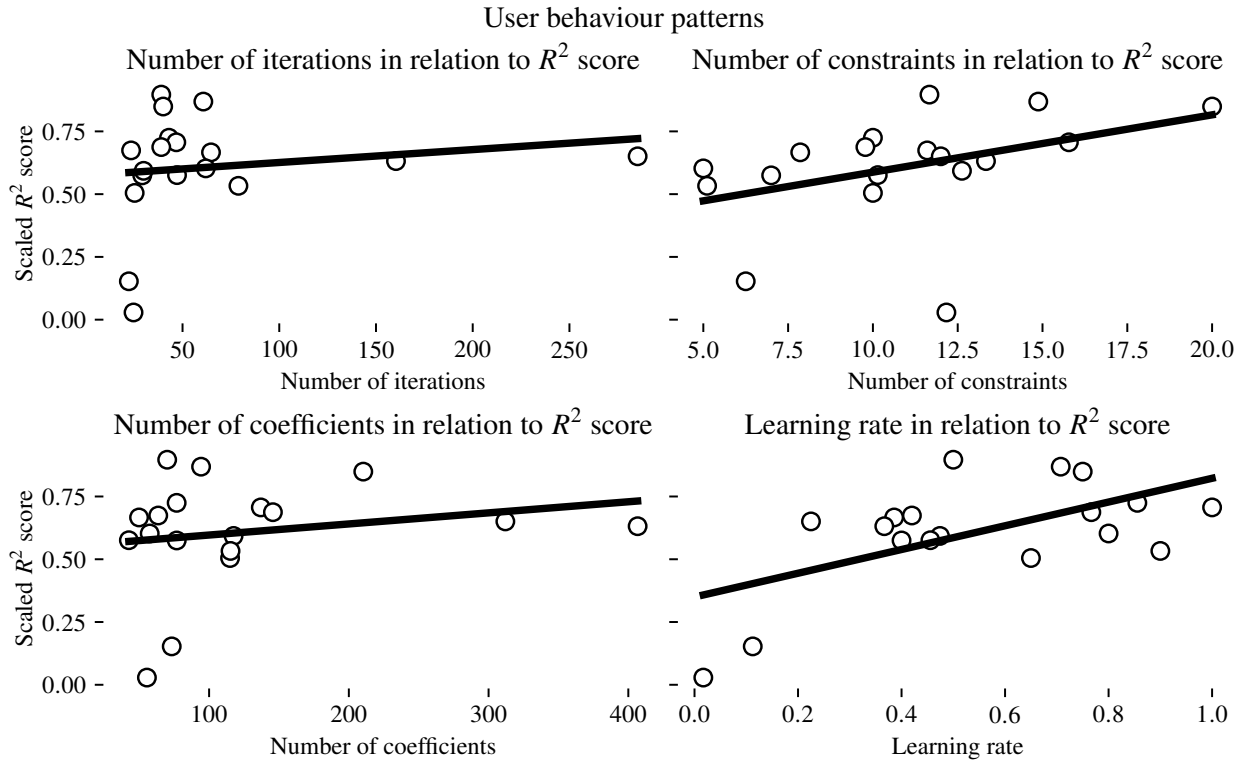}}
    \caption{Average model properties in relation to the average %
performance for each participant. A solid line represents a positively sloped line of best fit.}
    \label{fig:user_behave}
\end{figure}

We also studied those relations on the level of the users. We calculated the average properties and the normalized average performance of all the models built by each user and scattered them against each other. These plots together with their lines of best fit are presented in Figure \ref{fig:user_behave}. Here we can again see a positive correlation between the performance and the learning rate as well as the number of constraints $l$. Table \ref{tab:user_stats} reveals that the top five performers employed higher learning rates and more constraints in their models. In addition, they frequently adjusted hyperparameters, suggesting that this tuning, along with higher learning rates and more constraints, is associated with better performance.

\begin{table}[!ht]
    \centering
    \caption{Use statistics for all users, the five users with the best performance and the five users with the worst performance. For each group, the average number of iterations of model building, the average rate of reset operations, the average rate of hyperparameter changes, the average learning rate and, the average number of constraints $l$ are compared with each other.}
    \begin{tabular}{r|c|c|c}

        & All users & Top 5 users & Bottom 5 users \\ 
        \hline\hline
        Number of Iterations & 52.9919(51.3835) & 44.7111(5.7157) & 28.4008(9.8094) \\ \hline
        Rate of resets & 0.0901(0.0631) & 0.1308(0.0816) & 0.0785(0.0435) \\ \hline
        Rate of new parameters & 0.0867(0.1038) & 0.1444(0.1565) & 0.0673(0.0466) \\ \hline
        Learning Rate & 0.5421(0.2714) & 0.8113(0.1090) & 0.2828(0.1598) \\ \hline
        Number of constraints & 11.0382(3.7507) & 14.1333(3.8505) & 9.5008(2.4914)
    \end{tabular}
    \label{tab:user_stats}
\end{table}

\section{Conclusion}
\label{conclusion}
We presented HuGuR, an interactive and transparent human-in-the-loop approach to
regression with structured features. The framework was instantiated for the novel task of permutation regression, where the features are essentially constraints on the orderings of items. 
We conducted an user study with 21 participants to measure HuGuR's performance in its designated field of application, the interactive model generation by a variety of different users. The performance of human-generated models was compared to the one of multiple baseline approaches.

As shown in Section \ref{experiments}, HuGuR outperforms all other methods, most of them significantly, on small data sets and even performs significantly better than the other methods, except for the the sequence encoding based approach, over all data sets. While HuGuR is not significantly better than the sequence encoding based approach, it is also not significantly worse. Thus it performs on par with the most complex neural network based comparison method.

In future work we want to study differences in the amount of trust users have in machine learning models built by themselves using the HuGuR interface in contrast to conventional machine learning models built in a more black-box way. Additionally, we want to investigate how we can adapt our approach to handle other closely related types of data, such as strings, sequences, trees, or graphs.

\bibliographystyle{unsrt}  
\bibliography{my_bib}

\end{document}